\pdfoutput=1

\documentclass[11pt]{article}

\usepackage{acl}

\usepackage{amsmath,amsfonts,bm}

\def\eqref#1{equation~\ref{#1}}

\def\1{\bm{1}}

\DeclareMathAlphabet{\mathsfit}{\encodingdefault}{\sfdefault}{m}{sl}
\SetMathAlphabet{\mathsfit}{bold}{\encodingdefault}{\sfdefault}{bx}{n}

\usepackage{times}
\usepackage{latexsym}
\usepackage{url}
\usepackage{graphicx}
\usepackage{booktabs}
\usepackage{algorithm}
\usepackage{algpseudocode}
\usepackage{multirow}
\usepackage{enumitem}
\usepackage{cleveref}
\usepackage{xcolor,colortbl}
\usepackage{wrapfig,lipsum,booktabs}
\usepackage{bbm}
\usepackage{subfigure}

\algnewcommand\algorithmicforeach{\textbf{for each}}
\algdef{S}[FOR]{ForEach}[1]{\algorithmicforeach\ #1\ \algorithmicdo}

\usepackage[T1]{fontenc}

\usepackage[utf8]{inputenc}

\usepackage{microtype}

\usepackage{inconsolata}

\makeatletter
\newcommand*{\rom}[1]{\expandafter\@slowromancap\romannumeral #1@}
\makeatother
\newcommand{\method}[1]{\textsc{Branch-Solve-Merge}}
\newcommand{\methodsmall}[1]{BSM}

\title{Branch-Solve-Merge Improves Large Language Model \\ Evaluation and Generation}

\author{Swarnadeep Saha\footnotemark[1] \\ UNC Chapel Hill\\
\And Omer Levy \\ Meta \\
\And Asli Celikyilmaz \\ Meta \\
\AND Mohit Bansal \\ UNC Chapel Hill\\
\And ~~Jason Weston \\ ~~Meta \\
\And ~Xian Li \\ ~Meta \\
}

\begin{document}
\maketitle
\renewcommand{\thefootnote}{\fnsymbol{footnote}}
\footnotetext[1]{Work done during an internship at Meta.}
\begin{abstract}
Large Language Models (LLMs) are frequently used for multi-faceted language generation and evaluation tasks that involve satisfying intricate user constraints or taking into account multiple aspects and criteria. However, their performance can fall short, due to the model's lack of coherence and inability to plan and decompose the problem. We propose \method{} (\methodsmall{}), a Large Language Model program \citep{schlag2023large} for tackling such challenging natural language tasks. It consists of \emph{branch}, \emph{solve}, and \emph{merge} modules that are parameterized with specific prompts to the base LLM. These three modules plan a decomposition of the task into multiple parallel sub-tasks, independently solve them, and fuse the solutions to the sub-tasks. We apply our method to the tasks of LLM response evaluation and constrained text generation and evaluate its effectiveness with multiple LLMs, including Vicuna, LLaMA-2-chat, and GPT-4. BSM improves the evaluation correctness and consistency for each LLM by enhancing human-LLM agreement by up to 26\%, reducing length and pairwise position biases by up to 50\%, and allowing LLaMA-2-chat to match or outperform GPT-4 on most domains. On a constraint story generation task, BSM improves the coherence of stories while also improving constraint satisfaction by 12\%.
\end{abstract}

\section{Introduction}

\begin{figure*}[t]
    \centering
    \includegraphics[width=\textwidth]{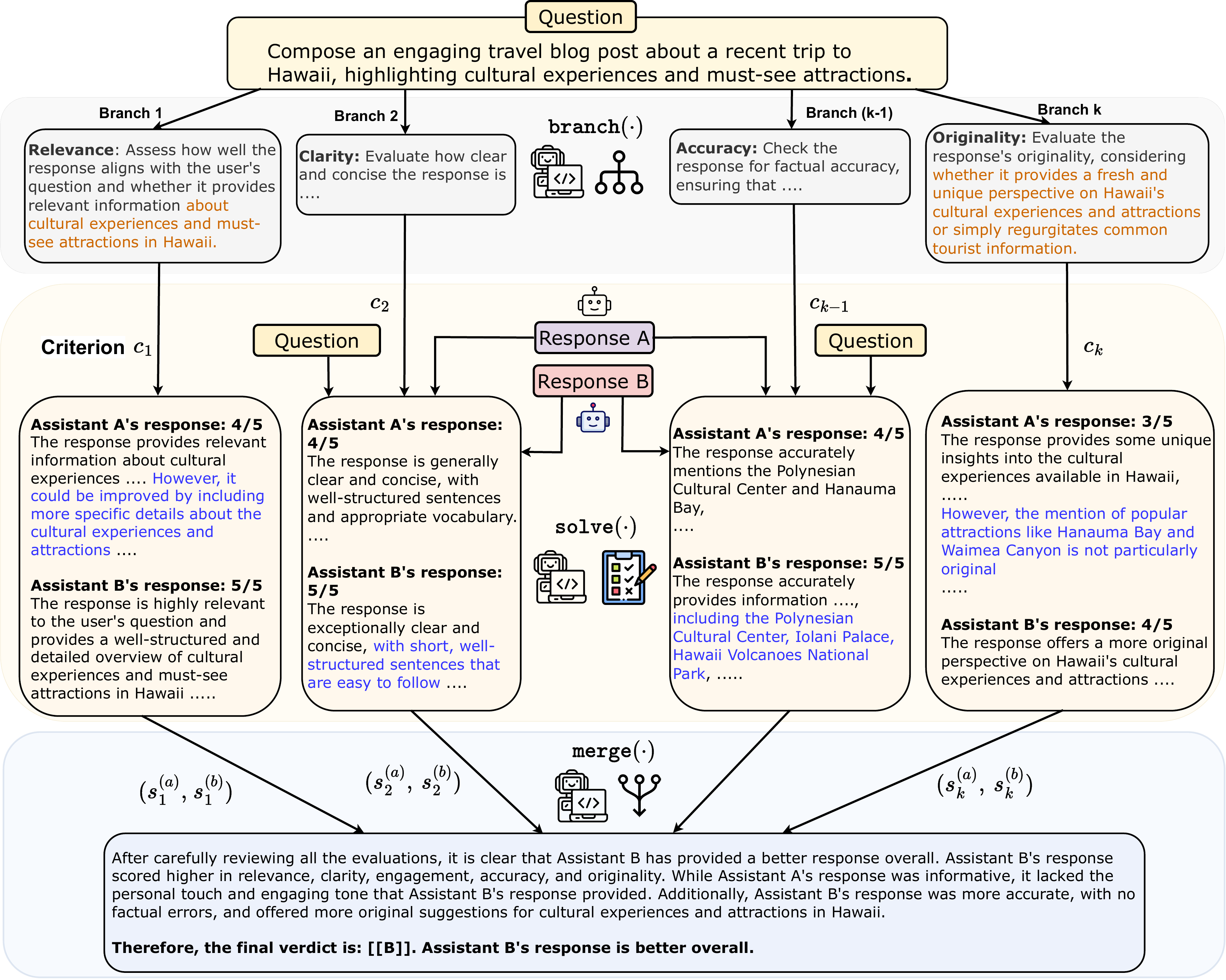}
    \vspace{-5pt}
    \caption{An illustration of \method{} with LLaMA-2-70B-chat for pairwise evaluation of LLM responses. Given a question and two LLM responses A and B, BSM generates a preference judgment. The Branch module conditions on the question to generate a question-specific evaluation plan which in this case consists of different criteria like `Relevance' to the Hawaii trip topic, `Clarity', etc. The `Solve' module evaluates the response pairs for each criteria (branch) independently and the `Merge' module combines the individual judgments to generate the final verdict, in this case that B is the better response.}
    \label{fig:intro}
\end{figure*}

Large Language Models (LLMs) are widely used for various text generation tasks~\citep{radford2019language, brown2020language, openai2023gpt4, chowdhery2022palm, touvron2023llama}. It has also become common to employ them as evaluators of such LLM generations in order to assess, critique and improve the outputs~\citep{zheng2023judging, bai2022constitutional}. However, LLMs still struggle with tasks that have intricate requirements like satisfying a set of constraints or meeting objectives that are, in general, multi-dimensional (e.g., evaluating the quality of generated text against certain diverse criteria). This appears to primarily stem from the model's lack of self-consistency and inability to plan~\citep{yao2023tree, bubeck2023sparks}. 
Recent research has tried to mitigate these limitations by developing iterative methods that involve eliciting reasoning, planning, and refinement, but so far they are still considered as open problems~\citep{bai2022constitutional, madaan2023self, ganguli2023capacity, yao2023beyond, chen2023teaching, li2023self, huang2023large}.

In this work, we propose \method{} (\methodsmall{}), 
a decomposition method for solving such multi-faceted natural language tasks. Our approach is an instance of a \textit{Large Language Model program}~\citep{schlag2023large, dohan2022language} and consists of three modules: \emph{branch}, \emph{solve}, and \emph{merge} that are parameterized with specific prompts to an underlying LLM. Given an arbitrary user task, the `branch' module generates a solution plan by decomposing the task into multiple \emph{parallel} sub-tasks, where each sub-task is represented by a unique branch, representing different components required to solve the overall problem. The `solve' module then solves each of these independent sub-problems. Finally, the `merge' module fuses the solutions to these sub-problems to generate the overall solution. We apply our method to two challenging tasks where LLMs are commonly utilized but their performance still lags behind humans:
\begin{itemize}[nosep, wide=0pt, leftmargin=*, after=\strut]
    \item \textbf{Evaluation of LLM Outputs~\citep{zheng2023judging}.} 
    LLMs are now regularly used to perform automatic evaluation of model responses, e.g., to user queries~\citep{dubois2023alpacafarm}.  Evaluating LLMs holistically is challenging because of their ability to generate long-form answers to arbitrary user questions~\citep{zheng2023judging}, the lack of reliability originating from many biases~\citep{zheng2023judging, wu2023style, wang2023large}, and reliance on hand-designed evaluation plans that impact the method's generalization, introducing unintended human biases~\citep{liu2023gpteval, wu2023style}. BSM can be applied to this task by each branch assessing different aspects and criteria that require evaluation.\footnote{Subsequently, we will refer to the task as `LLM Evaluation'. In the scope of our study, this will involve the pairwise evaluation of the response quality of two LLM outputs.}
    \item \textbf{Constrained Text Generation.} State-of-the-art LLMs struggle with constrained text generation tasks, e.g., the constraint of writing a story that should include several concepts. Models commonly either violate constraints, or else generate text that is incoherent in order to satisfy these constraints~\citep{bubeck2023sparks, yao2023collie}. BSM can be applied to this task by each branch writing part of the story satisfying only some of the constraints, followed by a final merge.
\end{itemize}

We apply \methodsmall{} to both these problems, 
see Fig.~\ref{fig:intro} and Fig.~\ref{fig:constrained_gen},
and evaluate its effectiveness with multiple open-source and black-box LLMs of varying sizes and strengths including LLaMA-2-7B-chat~\citep{touvron2023llama}, Vicuna-33B~\citep{vicuna2023}, LLaMA-2-70B-chat, and GPT-4~\citep{openai2023gpt4}. BSM significantly improves both tasks, addressing the aforementioned limitations of LLM evaluation and generation:
\begin{itemize}[nosep, wide=0pt, leftmargin=*, after=\strut]
\item BSM improves \emph{correctness} of LLM evaluation. In particular, on the MT-Bench benchmark~\citep{zheng2023judging}, BSM improves LLM-human agreement for evaluating multi-turn questions belonging to different domains including writing, coding, reasoning, and mathematics. For example, compared to zero-shot prompting and self-consistency~\citep{wang2022self} baselines, BSM with LLaMA-2-70B-chat improves LLM-human agreement by up to absolute 26\% and even matches or outperforms GPT-4 on many domains. BSM with GPT-4 improves agreement by a further 3\% over GPT-4. Overall, these findings point to BSM's ability to evaluate LLM responses to arbitrary user questions from diverse domains and to improve any base LLM as an evaluator.
\item  BSM also improves the \emph{consistency} of LLM evaluation. It significantly reduces position, length, and self-enhancement biases of LLM-based evaluators. For instance, BSM with LLaMA-2-70B-chat reduces position bias by up to absolute 50\%. Importantly, BSM with GPT-4 also improves GPT-4's reliability as an evaluator when evaluating its own responses. 

\item For the constrained story generation task, BSM generates more coherent stories, which are preferred by a GPT-4 judge a substantial 93\% of the time compared to a zero-shot baseline. It also improves constraint satisfaction by 12\%.
\end{itemize}

Overall, \methodsmall{} provides a framework for planning and task decomposition for addressing challenging multi-faceted language generation and evaluation tasks. As the approach is framed as a generic LLM program, it can be applied to any underlying LM and potentially a wide range of tasks.

\section{Related Work}

\noindent \textbf{LLM Programs and Decomposing Complex Tasks.} 
LLM programs such as \methodsmall{} solve complex problems with an algorithm that breaks the problem down into multiple steps and each step is then parameterized with a different prompt to an underlying LLM~\citep{schlag2023large, dohan2022language, creswell2022faithful}.
Complex tasks, in general, require task decomposition~\citep{khot2022decomposed} and planning~\citep{yao2022react, huang2022language,yao2023tree, ning2023skeleton}. This has motivated a lot of recent work on advanced prompting methods~\citep{khot2022decomposed, zhou2022least, wang2023plan, dua2022successive, saha2022summarization, saha-etal-2023-murmur, khot2021text, gupta2023visual, cho2023visual}. However, most of these works typically focus on reasoning problems (like commonsense, symbolic, or mathematical) that benefit from \emph{sequential} decompositions. We, however, study tasks that benefit from branching into \emph{parallel} decompositions, in particular LLM evaluation and constrained text generation. As well as being an LLM program, BSM is also an instance of Graph-of-Thoughts (GoT) prompting~\citep{lei2023boosting, besta2023graph} because the execution trace takes the shape of a graph. GoT defines a wide array of LLM programs, including refining, backtracking and skipping graph nodes, which we do not consider here. Our work develops a specific fixed program, and applies it to the challenging tasks of evaluating or improving language models. 
\paragraph{LLM Evaluation.} A fundamental challenge with the rapid progress of LLMs is evaluating their capabilities holistically~\citep{chang2023survey, liang2022holistic}. Human evaluation is difficult and expensive~\citep{smith2022human}.
On the other hand, LLMs, by being trained with RLHF, are shown to exhibit alignment with humans~\citep{ouyang2022training, bai2022training}. Hence, a standard procedure for comparing and evaluating LLM generations is by utilizing a strong LLM like GPT-4~\citep{bubeck2023sparks, openai2023evals, dubois2023alpacafarm, zhou2023lima, chiang2023can, wang2023far, hada2023large, liu2023gpteval} on different benchmarks~\citep{zhong2023agieval, kopf2023openassistant, zheng2023judging}. LLM-based evaluators are not fair evaluators~\citep{wang2023large, wu2023style} and there have been proposals of using multi-agent debate~\citep{chan2023chateval} or developing wider and deeper LLMs~\citep{zhang2023wider}. In contrast, \methodsmall{} improves LLM evaluation through an intuitive and general decomposition-based approach that can be applied on top of any LLM to evaluate responses for a wide range of tasks.

\paragraph{Constrained Text Generation.} Recent works evaluate LLMs for their capabilities in the more difficult setting of controllable and constrained text generation~\citep{keskar2019ctrl, dathathri2019plug, lu2021neurologic, lu2022neurologic, lin2020commongen, li2022diffusion} and show that even GPT-4 struggles with such planning-based tasks~\citep{bubeck2023sparks, madaan2023self, yao2023collie}. We experiment with such a constrained story generation task and show the promise of \methodsmall{}.

\section{\method{}}

We first introduce some notation to formally describe \methodsmall{}. Let $p_\theta$ denote an LLM  with parameters $\theta$. We also denote $x = x_{1, \cdot \cdot \cdot, n}$ as a sequence of $n$ tokens, such that $p_\theta(x) = \prod_{i=1}^n p_\theta (x_i|x_{1,\cdot \cdot \cdot,i-1})$.
\methodsmall{} is an LLM  program that aims to solve complex planning-based tasks with three neural modules: branch, solve, and merge. Each module is parameterized with unique prompts to the LLM $p_\theta$. The LLM program further defines an algorithm on top of these modules, acting as a controller and invoking a module at each step of the algorithm.

\subsection{Components of \method{}}
\paragraph{LLM Program.} For a given task, \methodsmall{} defines a controller as an algorithm that lays out the transition logic between the modules. Let us denote the three modules with their functional forms: $\texttt{branch}(\cdot)$, $\texttt{solve}(\cdot)$, and $\texttt{merge}(\cdot)$. Then the program is defined as $\texttt{Prog}: (x, \texttt{branch}(\cdot), \texttt{solve}(\cdot), \texttt{merge}(\cdot)) \rightarrow y$, taking as input a task instance $x$, along with the module implementations and generating an output $y$.

\paragraph{Branch Module.} Given a task, the branch module generates multiple sub-tasks where each sub-task is represented by a unique branch. Branching into sub-problems allows task decomposition such that each part can be solved independently in parallel, at which point the partial solutions are combined. Formally, given a task input $x$, we define a `branch' prompt $\texttt{prompt}_{\texttt{branch}}(x)$ that can be wrapped around $x$ with branching instructions and some demonstrations (if available). Conditioning on the prompt, the LLM $p_\theta$ generates a set of $k$ sub-problems $X = \{x^{(1)}, x^{(2)}, \cdot \cdot \cdot, x^{(k)}\}$, where $k$ is referred to as the branching factor. The sub-problems are generated auto-regressively as a sequence of tokens: $X \sim p_\theta(X | \texttt{prompt}_{\texttt{branch}}(x))$. Importantly, the flexibility of our method comes from the fact that for a given problem, the LLM itself decides (generates) the sub-problems and the corresponding branching factor. 

\paragraph{Solve Module.} The solve module solves the task at hand by generating an output $y^{(i)}$ for a branch task input $x^{(i)}$. Similar to the branch prompt, we define a `solve' prompt $\texttt{prompt}_{\texttt{solve}}(x^{(i)})$, conditioning on which the LLM generates a solution $y^{(i)} \sim p_\theta(y^{(i)} | \texttt{prompt}_{\texttt{solve}}(x^{(i)}))$ for each branch.

\paragraph{Merge Module.} The merge module fuses the solutions to the sub-problems to generate a global solution to the main problem. This is done through a `merge' prompt $\texttt{prompt}_{\texttt{merge}}(Y)$ that generates a merged solution $y \sim p_\theta(y | \texttt{prompt}_{\texttt{merge}}(Y))$, conditioning on a set of sub-solutions $Y = \{y^{(1)}, y^{(2)}, \cdot \cdot \cdot, y^{(k)}\}$. Conceptually, the merge module learns an aggregator function that could aggregate a set of values (using an aggregation operator) or fuse pieces of text, depending on the task. 

Next, we motivate and conduct case studies of BSM with two challenging NLP tasks: LLM evaluation and constrained generation.

\subsection{\methodsmall{}: Case Study with LLM Evaluation}

\paragraph{Task Description.} We consider the task of evaluating LLM-based chat assistants. Formally, given an open-ended question and a pair of responses from two LLM agents, the task requires producing a preference judgement of which response is better or if it is a tie (see Fig.~\ref{fig:intro}). Evaluating LLM responses is challenging for many reasons: 

\begin{enumerate}[nosep, wide=0pt, leftmargin=*, after=\strut]
\item \textbf{Long-form answers to arbitrary questions.} With the goal of providing a general-purpose assistant, the user asks arbitrary questions from any domain, and the LLM responds with  long-form answers~\citep{zheng2023judging}. Based on the initial model response, the user can ask follow-up questions. Depending on the type of question, the evaluation process must consider the intent of the question, what is expected from an ideal response, and what criteria to evaluate on. 

\item \textbf{LLM evaluators are prone to biases.} LLM-based evaluators are not reliable and are prone to different biases including (a) \emph{Position Bias}: evaluation changes based on the encoding order of the responses, (b) \emph{Length Bias}: tendency to favor longer responses, (c) \emph{Self-enhancement Bias}: the LLM-evaluator favoring its own responses~\citep{zheng2023judging}.

\item \textbf{GPT-4 as evaluator is expensive.} While API-based models like GPT-4 are fairly good evaluators~\citep{zheng2023judging}, these models are proprietary and charge users per token generated. Current open-source alternatives correlate less well with humans and are much more susceptible to the aforementioned biases. 

\item \textbf{Hand-designing evaluation plans is not scalable.} A robust evaluator should generalize well, capable of evaluating responses to arbitrary questions and hence, hand-designing the evaluation plan for every task is not desirable~\citep{liu2023gpteval}. For example, see Fig.~\ref{fig:intro}, where evaluating responses to a `writing' question requires considering factors like `Relevance', `Clarity', etc whereas if the question is a `coding' question (see Fig.~\ref{fig:branch_turn2_code} in the Appendix), one should evaluate for `Code Correctness', `Code Readability', etc. 
\end{enumerate}

Hence, given the multi-faceted nature of this evaluation task, we develop a version of \methodsmall{}, as described below. For this study, we focus on evaluating two-turn conversational questions although our method is generally applicable for any number of turns. Let us denote the first question as $q_1$ and the follow-up question as $q_2$. Let the responses from the two LLMs $A$ and $B$ be $r_1^{(A)}$ and $r_1^{(B)}$ for $q_1$, and $r_2^{(A)}$ and $r_2^{(B)}$ for $q_2$. 

\paragraph{Branch Module for LLM Evaluation.} It generates an evaluation plan i.e., a set of evaluation criteria that the response will be evaluated against. The branch module only conditions on the input question and for turn-1 questions, is defined as as \texttt{branch($q_1$)}, while for turn-2 questions, it conditions on both turn-1 and turn-2 questions, represented as \texttt{branch($q_1, q_2$)}. The output is a set of evaluation criteria, $\texttt{branch}(q) \rightarrow \{c_i\}_{i=1}^k$, where each $c_i$ is the title of the criterion (e.g., `Relevance') and a short description of how to evaluate for it (e.g., `Assess how well the response aligns with the user's question ... and must-see attractions in Hawaii.'). See Fig.~\ref{fig:intro} and~\ref{fig:branch_turn2_code} for examples of generated branches for different questions.

\paragraph{Solve Module for LLM Evaluation.} It compares and evaluates the responses based on a  specific criterion. The output of the evaluation is a pair of scores (within a specified range, according to the solving instruction, e.g., 1-5) for each of the responses. For example, given an evaluation criterion $c$, we denote the solve module for a question $q$
as: $\texttt{solve}(q, r_1^{(A)}, r_1^{(B)}, c) \rightarrow (s^{(A)}, s^{(B)})$,
where $s^{(A)}$ and $s^{(B)}$ are the evaluation scores assigned to the two assistant responses.
Note that the solve module is not symmetric i.e., the encoding order of the two responses is important (and we address this below in our LLM program). The module additionally generates explanations along with the scores. Fig.~\ref{fig:intro} shows example generations from the solve module with a LLaMA-2-70B-chat model.

\paragraph{Merge Module for LLM Evaluation.} We develop two variants of the merge module. A simple non-neural variant sums up the scores across all branches. We also develop a neural LLM variant that conditions on the individual evaluations and generates the final verdict with a model-decided aggregation strategy, denoted as $\texttt{merge}(q, \{c_i\}_{i=1}^k, \{s_i^{(A)}\}_{i=1}^k,  \{s_i^{(B)}\}_{i=1}^k) \rightarrow y$, where the evaluation criteria $\{c_i\}_{i=1}^k$ are the outputs of the branch module and $s_i^{(A)}$ and $s_i^{(B)}$ are the criterion-wise evaluations (scores and explanations) of the two assistant responses generated from the solve module. The final verdict is $y \in \{A, B, tie\}$.

\paragraph{LLM Program for LLM Evaluation.} The overall LLM program pseudocode is given in Algorithm~\ref{algo:eval}. To account for position bias, the program executes two independent runs of BSM by swapping the encoding order of the responses in the `solve' module. The final judgment is either `A' or `B' if and only if the judgement is consistent for both orders, otherwise it is a `tie'.

\subsection{\methodsmall{}: Case Study with Constrained Gen}

\paragraph{Task Description.} Our next case study shows the general applicability of BSM by applying it to a completely different task, that of LLM generation. We consider a constrained story generation task -- given a set of concepts $l$, the task is to generate a coherent story $y$ by including all concepts in it (see Fig.~\ref{fig:constrained_gen} in the Appendix). When the number of concepts is large, LLMs tend to either leave out some concepts or generate text that is incoherent. The task requires composition incorporating the various constraints. 

\paragraph{Branch Module for Constrained Generation.} The branch module $\texttt{branch}(l) \rightarrow (l_1, l_2, t)$ proposes a story generation plan, consisting of (1) two subsets of concepts $l_1$ and $l_2$ and (2) a story topic $t$. The two subsets represent sub-problems of the original task with a smaller number of concepts. The story topic ensures that all sub-stories generated as part of BSM belong to the same topic. 

\paragraph{Solve Module for Constrained Generation.} The solve module $\texttt{solve}(l_i, t) \rightarrow y_i$ conditions on a subset of concepts $l_i$ and the story topic $t$ to generate a story $y_i$ on that topic, while also including all concepts in $l_i$. Intuitively, `solving' the constrained generation task is easier with a smaller number of concepts.

\paragraph{Merge Module for Constrained Generation.} The merge module $\texttt{merge}(y_1, y_2) \rightarrow y$ conditions on two intermediate stories and fuses them together to generate the final story $y$. Since both intermediate stories belong to the same high-level topic, the fusion can lead to a final coherent story. Overall, BSM ensures better constraint satisfaction by solving sub-problems and maintains coherency through a top-level plan that includes a story topic.

\section{Experiments}

\subsection{Large Language Model Evaluation}
\label{sec:exp_eval}
\subsubsection{Experimental Setup} 
\paragraph{Dataset.} We experiment with the MT-Bench dataset, that evaluates LLMs as judges of other LLM's responses when acting as helpful AI assistants in multi-turn conversations~\citep{zheng2023judging}. It consists of instructions from 8 diverse domains e.g., writing, reasoning, math, coding, etc.

\paragraph{Evaluation Metrics.} We evaluate BSM (and baselines) using the following four metrics.

\begin{itemize}[nosep, wide=0pt, leftmargin=*, after=\strut]
\item \textbf{LLM-Human Agreement (Ag).} Following past work~\citep{zheng2023judging}, we report LLM-human agreement $\in [0, 1]$ individually for turn-1 and turn-2 questions, and their combination.

\item \textbf{Position Bias (PB).} To evaluate whether BSM helps reduce the consistency problem with LLM-based evaluators, we report PB, which is the fraction of samples where the judgment changes based on the encoding order of the responses.
\item \textbf{Length Bias (LB).} We measure LB as the fraction of samples where humans prefer the shorter response but the evaluator model does not. In other words, we compute how often an evaluator chooses the longer response when according to human preference, it should not.
\item \textbf{Self-enhancement Bias (SB).} SB refers to an evaluator model preferring its own responses. Evaluating this bias in isolation is challenging because knowing when the model is choosing its own response because of this bias and not for another reason is an interpretability question. However, the question we are interested in studying here is the following: \textit{When an LLM is evaluating its own responses (which is a common phenomenon when using LLMs as evaluators), does BSM lead to better and more reliable evaluation?}
We measure this by considering the following setting. We use GPT-4 as the base judge model and consider the subset of samples from the MT-Bench benchmark where one of the responses is \emph{also} generated by GPT-4. If BSM with GPT-4 improves human-agreement for this subset of samples, it suggests that \emph{even} in scenarios where model A is judging its own outputs, BSM (with model A) leads to a better evaluator. While this does not necessarily compute whether an evaluator has less SB, it does verify whether the evaluator model correlates better with humans even when it is evaluating its own responses.
\end{itemize}

While multiple past works have highlighted the importance of these biases~\citep{zheng2023judging, wu2023style}, we measure all of them with concrete metrics within the same evaluation framework. Conceptually, `Ag' evaluates \emph{correctness} while `PB' for example evaluates \emph{consistency} of LLM-based evaluators. These are complementary aspects and an ideal evaluator should perform well in all metrics for it to be reliably used.

\begin{table*}[]
\centering
\small
\begin{tabular}{lccc|cccccc}
\toprule
        \multirow{2}{*}{Method} & \multicolumn{3}{c|}{Overall} & \multicolumn{3}{c}{Turn-1}                                        & \multicolumn{3}{c}{Turn-2}                                         \\ \cmidrule(lr){2-4} \cmidrule(lr){5-7} \cmidrule(lr){8-10}
        & Ag$\uparrow$ & PB$\downarrow$ & LB$\downarrow$ & Ag$\uparrow$ & PB$\downarrow$ & LB$\downarrow$ &
        Ag$\uparrow$ & PB$\downarrow$ & LB$\downarrow$ \\ \midrule
Zero-shot (Relative) & 0.43 & 51.66 & 54.88 & 0.53 & 42.66 & 50.00 & 0.34 & 60.66 & 59.42                               \\
Zero-shot (Absolute) & 0.45 & 30.00 & 48.87 & 0.56 & 19.33 & 43.75 & 0.34 & 40.66 & 53.62 \\
Plan\&Solve & 0.43 & 43.00 & 54.13 & 0.43 & 42.00 & 51.56 & 0.43 & 44.00 & 56.52  \\
Self-Consistency & 0.52 & 35.66 & 48.12 & 0.57 & 32.00 & 45.31 & 0.47 & 39.33 & 50.72 \\
BSM & \bf 0.55 & \bf 17.33 & \bf 39.09 & \bf 0.60 & \bf 14.66 & \bf 39.46 & \bf 0.50 & \bf 20.00 & \bf 39.13 \\
\bottomrule                     
\end{tabular}
\vspace{-5pt}
\caption{Comparison of zero-shot LLM evaluators (Relative and Absolute), Plan\&Solve, Self-Consistency, and BSM on the `writing' questions in the MT-Bench dataset. All methods use LLaMA-2-70B-chat as the base LLM. We report LLM-Human Agreement (Ag), Position Bias (PB), and Length Bias (LB) for turn-1 and turn-2 questions overall, and individually. \methodsmall{} improves agreement scores, and reduces position and length biases.
}
\vspace{-5pt}
\label{tab:writing_llama}
\end{table*}

\begin{table}[]
\centering
\small
\begin{tabular}{lccc}
\toprule
        & Ag$\uparrow$ & PB$\downarrow$ & LB$\downarrow$ \\ \midrule
Zero-shot (w/ GPT-4) & 0.51 & \bf 6.33 & 36.36 \\
BSM (w/ GPT-4) & \bf 0.54 & 7.33 & \bf 34.54 \\
\bottomrule
\end{tabular}
\caption{\label{tab:self-enhancement bias} BSM leads to less self-enhancement bias. BSM obtains better agreement for the fraction of samples where one of the responses is also generated by GPT-4.}
\end{table}

\paragraph{Implementation Details.} We develop BSM on top of multiple LLMs of varying scales and capabilities: LLaMA-2-7B-chat, Vicuna-33B, LLaMA-2-70B-chat, and GPT-4. We implement all modules zero-shot, providing only module-specific instructions and assuming no access to demonstrations of how to branch, solve, or merge.

\paragraph{Baselines.} We compare our method, BSM, to (1) \textbf{two variants of zero-shot prompting with the same LLM}: a relative evaluator, that directly generates a preference judgment and an absolute evaluator, that generates two scores for the two responses and then the final preference is determined based on the higher score, (2) \textbf{plan\&solve prompting}~\citep{wang2023plan}, which plans (i.e., generates evaluation criteria) but instead of solving them independently, solves all branches together in one LLM call, (3) \textbf{self-consistency}~\citep{wang2022self}, which samples multiple evaluations from the prompted LLM (with temperature $0.7$) and chooses the majority vote as the final judgment. For fair comparison, self-consistency samples the same number of generations as the branching factor in BSM. We also note that self-consistency is a simple special case of BSM, where the branch module spawns multiple instances of the \emph{same} underlying problem (instead of sub-problems), solves them by sampling different solutions, and the merging operator is a majority vote. Refer to Appendix~\ref{appendix:exp_eval} for more details about the dataset, implementation, and baselines.

\subsubsection{Main Results}

\paragraph{\methodsmall{} improves LLM-human agreement and reduces biases.} Table~\ref{tab:writing_llama} evaluates the efficacy of \methodsmall{} with LLaMA-2-70B-chat as the base LLM, specifically focusing on the `writing' category of questions from the MT-Bench benchmark. We report our main findings below.

\begin{itemize}[nosep, wide=0pt, leftmargin=*, after=\strut]
\item \textbf{Overall agreement.} We find that BSM improves LLM-human agreement for both turn-1 and turn-2 questions, compared to all baselines. In particular, it obtains up to 12\% absolute improvement over Plan\&Solve, which specifically shows the utility of branching into and solving \emph{independent} sub-problems. BSM also outperforms Self-Consistency. As noted earlier, Self-Consistency is a special case of BSM. This result is noteworthy because both approaches leverage similar amounts of compute in generating multiple solutions -- but branching and solving the differing sub-problems provides superior results to solving the same problem multiple times.

\item \textbf{Turn-1 versus Turn-2 questions.} Evaluating turn-2 questions is harder because it requires additional contextualization of the responses for the turn-1 question. This is also reflected in all baseline methods (except for plan\&solve) exhibiting lower turn-2 agreement scores (e.g., zero-shot results drop from 0.53 in turn-1 to 0.34 in turn-2). BSM shows that a decomposition approach that generates an evaluation plan is particularly helpful for evaluating long context questions, resulting in more improvements for turn-2 questions (e.g., up to 16\% improvement). An illustration is shown in Fig.~\ref{fig:branch_turn2_code}, in which for the turn-2 question, the model generates `Adherence to Instructions' as the first criterion to evaluate.
\item \textbf{Position and Length Bias Reduction.} On top of improving LLM-human agreement, BSM helps reduce critical biases with LLM-based evaluators (e.g., up to 34\% reduction in PB). This is a direct consequence of BSM's task decomposition that helps reduce inconsistencies in evaluation. BSM's reduction in LB could be attributed to the following: when an evaluator branches into different criteria, if `length' is indeed one of the criteria that the responses should be evaluated against, it only counts as a single branch (i.e., one sub-problem) of the overall evaluation and hence, branching allows the model to explicitly evaluate for other criteria, beyond just length.
\item \textbf{Self-enhancement Bias Reduction.} Table~\ref{tab:self-enhancement bias} evaluates self-enhancement bias by comparing BSM (with zero-shot GPT-4) for the samples where one of the responses is also generated by GPT-4. We observe a 3\% better correlation with humans, suggesting that BSM improves evaluation even when the LLM judges its own outputs. 
\end{itemize}

\begin{table}[t]
\centering
\small
\resizebox{\columnwidth}{!}{%
\begin{tabular}{lccc}
\toprule
        Method & Ag$\uparrow$ & PB$\downarrow$ & LB$\downarrow$ \\ \midrule
        Zero-shot (w/ LLaMA-2-7B-chat) & 0.39 & 62.33 & 54.88 \\
BSM (w/ LLaMA-2-7B-chat) & \bf 0.41 & \bf 48.33 & \bf 53.38 \\ \midrule
Zero-shot (w/ Vicuna-33B) & 0.51 & 30.66 & 48.12 \\
BSM (w/ Vicuna-33B) & \bf 0.56 & \bf 20.00 & \bf 42.85 \\ \midrule
Zero-shot (w/ LLaMA-2-70B-chat) & 0.43 & 51.66 & 54.88                               \\
BSM (w/ LLaMA-2-70B-chat) & \bf 0.55 & \bf 17.33 & \bf 39.09 \\ \midrule
Zero-shot (w/ GPT-4) & 0.59 & 17.33 & 39.09 \\
BSM (w/ GPT-4) & \bf 0.62 & \bf 17.00 & \bf 36.84 \\
\bottomrule                     
\end{tabular}
}
\vspace{-5pt}
\caption{Comparison of zero-shot evaluation and BSM on `writing' questions with different base LLM evaluators. BSM improves agreement for all models and reduces biases for all models except GPT-4. 
}
\vspace{-10pt}
\label{tab:writing_all}
\end{table}

\begin{table*}[t]
\centering
\small
\begin{tabular}{lccc|ccc|ccc}
\toprule
        \multirow{2}{*}{Method} & \multicolumn{3}{c|}{Coding} & \multicolumn{3}{c|}{Reasoning}                                        & \multicolumn{3}{c}{Math}                                        \\ 
        \cmidrule(lr){2-4} \cmidrule(lr){5-7} \cmidrule(lr){8-10}
       & Ag$\uparrow$ & PB$\downarrow$ & LB$\downarrow$ & Ag$\uparrow$ & PB$\downarrow$ & LB$\downarrow$ & Ag$\uparrow$ & PB$\downarrow$ & LB$\downarrow$ \\ \midrule
Zero-shot (w/ LLaMA-2-70B-c) & 0.47 & 52.33 & 51.32 & 0.47 & 38.00 & 48.75 & 0.52 & 45.66 & 50.56\\
BSM (w/ LLaMA-2-70B-c) & \bf 0.61 & \bf 25.66 & \bf 42.47 & \bf 0.57 & \bf 20.33 & \bf 46.25 & \bf 0.64 & \bf 17.66 & \bf 34.83 \\
\color{gray} GPT-4   & \color{gray} \bf 0.61 & \color{gray} \bf 19.66 & \color{gray} \bf 38.93 & \color{gray} \bf 0.64 & \color{gray} 22.66 & \color{gray} 53.75 & \color{gray} 0.62 & \color{gray} 19.00 & \color{gray} 39.32 \\
\bottomrule                     
\end{tabular}
\vspace{-5pt}
\caption{Reference-based LLM evaluation for `Coding', `Reasoning', and `Math' question categories of MT-Bench. BSM improves reference-based evaluations and for math, outperforms GPT-4.}
\label{tab:eval_reference} 
\end{table*}

\begin{table*}[t]
\centering
\small
\begin{tabular}{llccc|cccccc}
\toprule
        \multirow{2}{*}{Domain} & \multirow{2}{*}{Method} & \multicolumn{3}{c|}{Overall} & \multicolumn{3}{c}{Turn-1}                                        & \multicolumn{3}{c}{Turn-2}                                        \\ \cmidrule(lr){3-5} \cmidrule(lr){6-8} \cmidrule(lr){9-11}
       & & Ag$\uparrow$ & PB$\downarrow$ & LB$\downarrow$ & Ag$\uparrow$ & PB$\downarrow$ & LB$\downarrow$ & Ag$\uparrow$ & PB$\downarrow$ & LB$\downarrow$ \\ \midrule
\multirow{3}{*}{Roleplay} & Zero-shot (w/ LLaMA-2-70B-c) & 0.55 & 29.66 & 51.67 &  0.61	& 30.00 & 48.14 & 0.50 & 29.33 & 55.88 \\
& BSM (w/ LLaMA-2-70B-c) & \bf 0.61 & \bf 11.00 & \bf 40.26 & \bf 0.66 &	\bf 10.66 & \bf 38.27 & \bf 0.56 & \bf 11.33 & \bf 42.64 \\
& \color{gray} GPT-4 & \color{gray} \bf 0.64 & \color{gray} 13.66 & \color{gray} 43.62 & \color{gray} 0.65 & \color{gray} 16.00 & \color{gray} 45.67 & \color{gray} \bf 0.63 & \color{gray} \bf 11.33 & \color{gray} \bf 41.17 \\ \midrule

\multirow{3}{*}{Extraction} & Zero-shot (w/ LLaMA-2-70B-c) &  0.40 & 70.66 & 51.82 & 0.46 & 61.33 & 51.47 & 0.33 & 80.00 & 52.08 \\
& BSM (w/ LLaMA-2-70B-c) & \bf 0.55 & \bf 31.33 & \bf 40.24 & \bf 0.55 & \bf 32.00 & \bf 45.58 & \bf 0.44 & \bf 30.66 & \bf 36.45 \\
& \color{gray} GPT-4 & \color{gray} \bf 0.71 & \color{gray} \bf 15.00 & \color{gray} \bf 33.53 & \color{gray} \bf 0.68 & \color{gray} \bf 13.33 & \color{gray} \bf 35.29 & \color{gray} \bf 0.75 & \color{gray} \bf 16.66 & \color{gray} \bf 32.29\\ \midrule

\multirow{3}{*}{Stem} & Zero-shot (w/ LLaMA-2-70B-c) & 0.46 & 59.33 & 55.31 & 0.50 & 52.66 & 51.19 & 0.43 & 66.00 & 61.40 \\
& BSM (w/ LLaMA-2-70B-c) & \bf 0.72 & \bf 10.33 & \bf 44.68 & \bf 0.70 & \bf 10.66  & \bf 40.47 & \bf 0.73 & \bf 10.00 & \bf 50.87 \\
& \color{gray} GPT-4 & \color{gray} \bf 0.72 & \color{gray} 13.66 & \color{gray} 46.80 & \color{gray} 0.68 & \color{gray} 16.66 & \color{gray} 44.04 & \color{gray} \bf \color{gray} 0.75 & \color{gray} 10.66 & \color{gray} \bf 50.87 \\ \midrule

\multirow{3}{*}{Humanities} & Zero-shot (w/ LLaMA-2-70B-c) & 0.46 & 59.00 & 45.69 & 0.51 & 52.00 & 49.18 & 0.41 & 66.00 & 43.33 \\
& BSM (w/ LLaMA-2-70B-c) & \bf 0.67 & \bf 18.00 & \bf 36.42 & \bf 0.63 & \bf 18.00 & \bf 39.34 & \bf 0.71 & \bf 18.00 & \bf 34.44 \\
& \color{gray} GPT-4 & \color{gray} \bf 0.73 & \color{gray} \bf 14.00 & \color{gray} 37.08 & \color{gray} \bf 0.70 & \color{gray} 19.33 & \color{gray} 42.62 & \color{gray} \bf 0.76 & \color{gray} \bf 8.66 & \color{gray} \bf 33.33 \\
\bottomrule                     
\end{tabular}
\caption{\label{tab:eval_no_reference}
 {LLM evaluation} for  `Roleplay', `Extraction', `Stem', and `Humanities' question categories of MT-Bench. 
We compare LLaMA-2-70B-chat BSM with the baseline zero-shot method, and also report GPT-4 results.  BSM obtains significant improvements over the LLaMA baseline, and matches or is close to GPT-4 agreement in three of the four domains, while sometimes outperforming GPT-4 in reducing biases.
}
\end{table*}

BSM not only leads to an improvement in overall LLM-human agreement (as per the `Ag' metric) but also on the fraction of samples where one response is generated by the same evaluator LLM (as per the `SB' metric), thus pointing to its robustness as an evaluation method. In summary, BSM improves both \emph{correctness} and \emph{consistency} of LLM-based evaluators.

\paragraph{BSM improves upon all zero-shot base LLMs.} We demonstrate the generalizability of BSM as an LLM program by implementing it on top of four different base LLMs, ranging from LLaMA-2-7B to GPT-4. As shown in Table~\ref{tab:writing_all}, BSM improves agreement with humans for all base LLMs, compared to a zero-shot baseline. Even though zero-shot GPT-4 is the state-of-the-art LLM-based evaluator, applying BSM obtains a further improvement of 3\%. Moreover, applying BSM to LLaMA-2-70B-chat makes it competitive with GPT-4 for turn-1 questions. BSM also significantly reduces position and length biases for all models except for GPT-4.

\paragraph{\methodsmall{} generalizes  to reference-based evaluations.}  
We find that BSM also excels at \emph{reference-based} evaluations for complex tasks like in math, reasoning, and coding~\citep{cobbe2021training, wei2022chain}. Following past work~\citep{zheng2023judging}, we evaluate responses for these categories by first generating an answer using GPT-4 and then appending it to the evaluation prompt, which is our baseline in this experiment.
For BSM, we then follow a similar recipe by conditioning the `solve' module on the GPT-4 generated answers. The key assumption here is that these answers are curated once and have limited variations unlike answers for open-ended questions. 
Table~\ref{tab:eval_reference} shows that \methodsmall{} significantly outperforms zero-shot baseline in all categories (by up to 14\% better agreement scores and 27\% better position bias in coding questions). On Math, it even outperforms the state-of-the-art GPT-4 evaluator, outperforming on all metrics.

\paragraph{\methodsmall{} generalizes across further domains.}
Table~\ref{tab:eval_no_reference} shows that BSM is capable of evaluating generations for questions in other categories like `Roleplay', `Extraction', `Stem', and `Humanities', with similar findings. 
See Appendix~\ref{appendix:eval_results} for details.

\paragraph{Scalability of \methodsmall{}'s branching.} One of the core strengths of BSM is its scalability -- it uses the same branch prompt (in Fig.~\ref{fig:branch_prompt}) for all evaluation domains (e.g., writing, code, reasoning, etc). The prompt only specifies the meaning of a branch for a given task and the LLM is capable of generating its own branches for different domains without any human intervention. We observe almost no overlap between the branch names of coding questions and writing questions. For example, the most occurring `writing branches' are \emph{Clarity, Relevance, Creativity, Accuracy, Engagement, Coherence, Originality, Completeness, Grammar and Readability, etc} whereas the most occurring `coding branches' are \emph{Efficiency, Completeness, Accuracy, Correctness, Code Readability, User Experience, Time Efficiency}. Branches for questions belonging to the same domain exhibit more overlap, as measured by the names of the branches. For example, `Correctness' is a branch that is generated for evaluating almost all coding problems; however, their descriptions are different and problem-specific (see Fig.~\ref{fig:branch_turn2_code} for an example).

\subsection{Constrained Text Generation} 
\label{sec:story_eval}

\subsubsection{Experimental Setup}

\paragraph{Dataset.} Our constrained story generation task is a more challenging variant of a generative commonsense reasoning task, CommonGen~\citep{lin2020commongen}. While the original task requires generating a single coherent sentence from 3 or 4 concepts, we increase the complexity of the task by having the model generate a concise story consisting of 10 concepts~\citep{madaan2023self}\footnote{\url{https://github.com/madaan/self-refine/blob/main/data/prompt/commongen/commongen_hard.jsonl}}. We experiment with 100 samples for the purpose of this study.

\paragraph{Evaluation Metrics.} We evaluate the generated stories along two axes: \textbf{constraints satisfaction} and \textbf{overall story quality}. For constraints satisfaction, we report two metrics: (a) \textbf{All Present (AP):} fraction of samples where all constraints are satisfied i.e., there are no missing concepts, and (b) \textbf{Missing Concepts (MC):} average percentage of missing concepts. Higher `all present' and lower `missing concepts' are preferable. We identify a concept as missing if it does not appear in the story in any word form.  For evaluating overall story quality, we conduct a pairwise evaluation with GPT-4. The evaluation prompt is provided in Fig.~\ref{fig:story_eval_prompt}. To account for position bias in this pairwise comparison, we follow our findings from the LLM Evaluation task and conduct each evaluation twice, by swapping the order of the stories and preferring one story over the other only if the evaluations are consistent.

\paragraph{Implementation Details.} We evaluate BSM using LLaMA-2-7B-chat and LLaMA-2-70B-chat. All modules generate text using greedy decoding. For the branch module, the LLM is prompted to divide the concepts into two groups.

\paragraph{Baselines.} We compare BSM to (1) \textbf{zero-shot prompting with the same LLM:} given a set of concepts, directly generates the story, (2) \textbf{plan\&solve prompting}, that first proposes a story topic (as a plan) and then generates a story on that topic, (3) \textbf{self-consistency}, where we first sample multiple stories and then prompt the LLM again to select one of the sampled stories that has more constraints satisfied.

\begin{table}[]
\centering
\small
\begin{tabular}{lcccc}
\toprule
                      \multirow{2}{*}{Method}  & \multicolumn{2}{c}{LLaMA-2-7B-chat}     & \multicolumn{2}{c}{LLaMA-2-70B-chat} \\ \cmidrule(lr){2-3} \cmidrule(lr){4-5}
                        & AP$\uparrow$ & MC$\downarrow$ & AP$\uparrow$ & MC$\downarrow$ \\ \midrule
Zero-shot  & 15.0  & 17.3               & 22.0 &  27.2                         \\
Plan\&Solve & 13.0 & 18.0 & 21.0 & 26.6 \\
Self-Consis & 19.0 & 16.6 & 24.0 & 20.1 \\
BSM       & \bf 23.0 &  \bf 15.5   &  \bf 28.0 & \bf 14.7           \\ 
\bottomrule            
\end{tabular}
\vspace{-5pt}
\caption{\label{tab:story} Constrained story generation evaluation results. BSM (with both LLaMA-2 models) improves constraint satisfaction in the generated stories.}
\vspace{-5pt}
\end{table}

\subsubsection{Results and Analysis}

\paragraph{Constraint Satisfaction.} Our main results are given in Table~\ref{tab:story}. They show that BSM with both model variants outperforms all baselines on our constraint satisfaction metrics. We also note that this is still a challenging task even for a stronger LLaMA-2-70B-chat model and the scale of the model has little impact on constraint satisfaction. For example, even BSM with LLaMA-2-70B-chat omits at least one concept for 72\% of the samples, echoing the findings from prior work that constrained text generation is hard even for state-of-the-art LLMs~\citep{yao2023collie}. We provide an analysis of missing concepts in BSM in Appendix~\ref{appendix:story_gen}.

\paragraph{Overall Story Quality.} BSM not only satisfies more constraints but almost always generates a more coherent story. We find that in a head-to-head comparison with the zero-shot prompting baseline (with LLaMA-2-70B-chat), stories generated by BSM are preferred a substantial 93\% of the time by GPT-4. This can be attributed to two aspects of BSM. First, in each of the branches, the model conditions on a lesser number of concepts and thus generates intermediate stories that by themselves are more coherent. Second, in the merging step, the model is able to condition on these two intermediate stories and generate a final story that further improves the coherence.

\section{Conclusion}

We presented BSM, an LLM program for improving LLM evaluation and generation. We conducted two case studies with different implementations of branch, solve, and merge modules, showcasing the effectiveness and generalizability of BSM. 

\section*{Limitations}

We list the limitations of our work below.
\begin{enumerate}
    \item Evaluating for safety, toxicity, and bias in LLM generations is also critical for a holistic evaluation of LLMs, however, we do not address this topic in our paper. 
    \item While BSM obtains improvements in length bias, we note that measuring length bias in isolation is
challenging because knowing whether the model prefers the longer response because of its length (and not for another reason) is an interpretability question and humans also tend to prefer longer
responses, especially for open-ended questions. 
\item Recursive or multi-level BSM, in which an LLM recursively branches into parallel sub-tasks is an interesting avenue for future work but we do not explore this in this work due to the increased computation cost. 
\item Decomposition into parallel sub-tasks should also help improve efficiency (e.g., compared to sequential decompositions)~\citep{ning2023skeleton} but in this work, we instead focused on improving the task performance.
\end{enumerate}

\section*{Acknowledgments}

The authors thank the reviewers for their helpful comments and suggestions.

\bibliography{custom}

\appendix

\begin{table}[t]
\centering
\small
\begin{tabular}{lccc}
\toprule
        & Ag$\uparrow$ & PB$\downarrow$ & LB$\downarrow$ \\ \midrule
Zero-shot (w/ LLaMA-2-70B-c) & 0.46 & 34.00 & 45.00 \\
BSM (w/ LLaMA-2-70B-c) & \bf 0.48 & \bf 23.77 & \bf 40.25 \\
\bottomrule
\end{tabular}
\caption{\label{tab:reasoning_without_ref} Results of reference-free evaluation of `Reasoning' questions. BSM outperforms the zero-shot baseline in evaluating `Reasoning' questions, even when reference answers are not used (on a random subset of 100 samples).}
\end{table}

\begin{table*}[t]
\centering
\small
\begin{tabular}{lccc|cccccc}
\toprule
        & \multicolumn{3}{c|}{Overall} & \multicolumn{3}{c}{Turn-1}                                        & \multicolumn{3}{c}{Turn-2}                                      \\ \cmidrule(lr){2-4} \cmidrule(lr){5-7} \cmidrule(lr){8-10}
        & Ag$\uparrow$ & PB$\downarrow$ & LB$\downarrow$ & Ag$\uparrow$ & PB$\downarrow$ & LB$\downarrow$ & Ag$\uparrow$ & PB$\downarrow$ & LB$\downarrow$ \\ \midrule
BSM & 0.55 & 17.33 & 39.09 & 0.60 & 14.66 & 39.46 & 0.50 & 20.00 & 39.13\\ 
BSM + SC & 0.55 & 15.33 & 39.09 & 0.61 & 10.66 & 39.06 & 0.49 & 20.00 & 39.13 \\
\bottomrule                     
\end{tabular}
\caption{\label{tab:bsm_sc} Effect of using Self-Consistency within each branch of \method{} (BSM+SC). Results are with the LLaMA-2-70B-chat model. While the overall agreement scores do not improve further, we obtain a further 2\% reduction in position bias.}
\end{table*}

\begin{table}[t]
\centering
\small
\begin{tabular}{lccc}
\toprule
        & Overall & Turn 1 & Turn 2 \\ \midrule
Vicuna-33B & 0.52 & 0.53 & 0.51 \\
BSM (w/ Vicuna-33B) & \bf 0.55 & \bf 0.56 & \bf 0.54 \\ \midrule
LLaMA-2-70B & 0.48 & 0.58 & 0.37                            \\
BSM (w/ LLaMA-2-70B) & \bf 0.53 & \bf 0.59 & \bf 0.47  \\ \midrule
GPT-4 & 0.61 & 0.59 & \bf 0.63 \\
BSM (w/ GPT-4) & \bf 0.63 & \bf 0.63 & 0.62 \\
\bottomrule                     
\end{tabular}
\caption{\label{tab:majority_vote} LLM-Human agreement scores for the `writing' category questions (overall and individually for turn-1 and turn-2). Here agreement is computed using majority voting (instead of treating each human vote per sample independently).}
\end{table}

\begin{table*}[t]
\small
\centering
\begin{tabular}{lccc|cccccc}
\toprule
        & \multicolumn{3}{c|}{Overall} & \multicolumn{3}{c}{Turn-1}                                        & \multicolumn{3}{c}{Turn-2}                                        \\ \cmidrule(lr){2-4} \cmidrule(lr){5-7} \cmidrule(lr){8-10}
   BF     & Ag$\uparrow$ & PB$\downarrow$ & LB$\downarrow$ & Ag$\uparrow$ & PB$\downarrow$ & LB$\downarrow$ & Ag$\uparrow$ & PB$\downarrow$ & LB$\downarrow$ \\ \midrule
2 & 0.50 & 22.00 & 49.20 & 0.49 & 24.00 & 45.00 & 0.50 & 22.00 & 56.52  \\
3 & 0.52 & 19.00 & 38.09 & 0.53 & 14.00 & 35.00 & 0.51 & 24.00 & 43.47 \\
4 & \bf 0.53 & 19.00 & 38.09 & 0.51 & 16.00 & 35.00 & 0.55 & 22.00 & 43.47 \\
5 & 0.52 & \bf 12.00 & \bf 34.92 & 0.51 & 12.00 & 35.00 & 0.54 & 12.00 & 34.78 \\
\bottomrule                     
\end{tabular}
\caption{\label{tab:branch} Impact of maximum Branching Factor (BF) on 100 samples of the `writing' category in LLM response evaluation with BSM on LLaMA-2-70B-chat.}
\end{table*}

\begin{table*}[t]
\small
\centering
\begin{tabular}{lccc|cccccc}
\toprule
        \multirow{2}{*}{Solving Technique} & \multicolumn{3}{c|}{Overall} & \multicolumn{3}{c}{Turn-1}                                        & \multicolumn{3}{c}{Turn-2}                                        \\ \cmidrule(lr){2-4} \cmidrule(lr){5-7} \cmidrule(lr){8-10}
        & Ag$\uparrow$ & PB$\downarrow$ & LB$\downarrow$ & Ag$\uparrow$ & PB$\downarrow$ & LB$\downarrow$ & Ag$\uparrow$ & PB$\downarrow$ & LB$\downarrow$ \\ \midrule
Eval Scale (1-5) & 0.52 & 12.00 & 34.92 & 0.51 & 12.00 & 35.00 & 0.54 & 12.00 & 34.78 \\
Eval Scale (1-10) & 0.50 & 18.00 & 36.50 & 0.51 & 18.00 & 40.00 & 0.50 & 18.00 & 30.43 \\
\bottomrule                     
\end{tabular}
\caption{Analysis of evaluation scale for LLM response evaluation, with 100 samples of `writing' category. BSM (with LLaMA-2-70B-chat) is fairly robust to such variations.}
\label{tab:solve} 
\end{table*}

\section{Additional Experiments: LLM Evaluation}

\label{appendix:exp_eval}
\subsection{Experimental Setup} 
\paragraph{Dataset.} We experiment with the MT-Bench dataset, that evaluates LLMs as judges of other LLM's responses when acting as helpful AI assistants in multi-turn conversations~\citep{zheng2023judging}. It consists of 2400 LLM responses and 3000 expert human judgements. LLM outputs are responses to 80 representative instructions from 8 diverse domains:  writing, roleplay, extraction, reasoning, math, coding, knowledge \rom{1} (STEM), and knowledge \rom{2} (humanities/social science). Each question is a conversational question, consisting of two turns, in which the turn-2 question is a follow-up to the turn-1 question. For each question, the dataset consists of responses from 6 different LLMs (Alpaca-13B, Vicuna-13b, LLaMA-13B, Claude-v1, GPT-3.5-turbo, and GPT-4), resulting in 15 possible response pairs.
Thus, the entire evaluation set consists of 300 response-pair samples per category.

\begin{figure*}[t]
\centering
  \includegraphics[width=\linewidth]{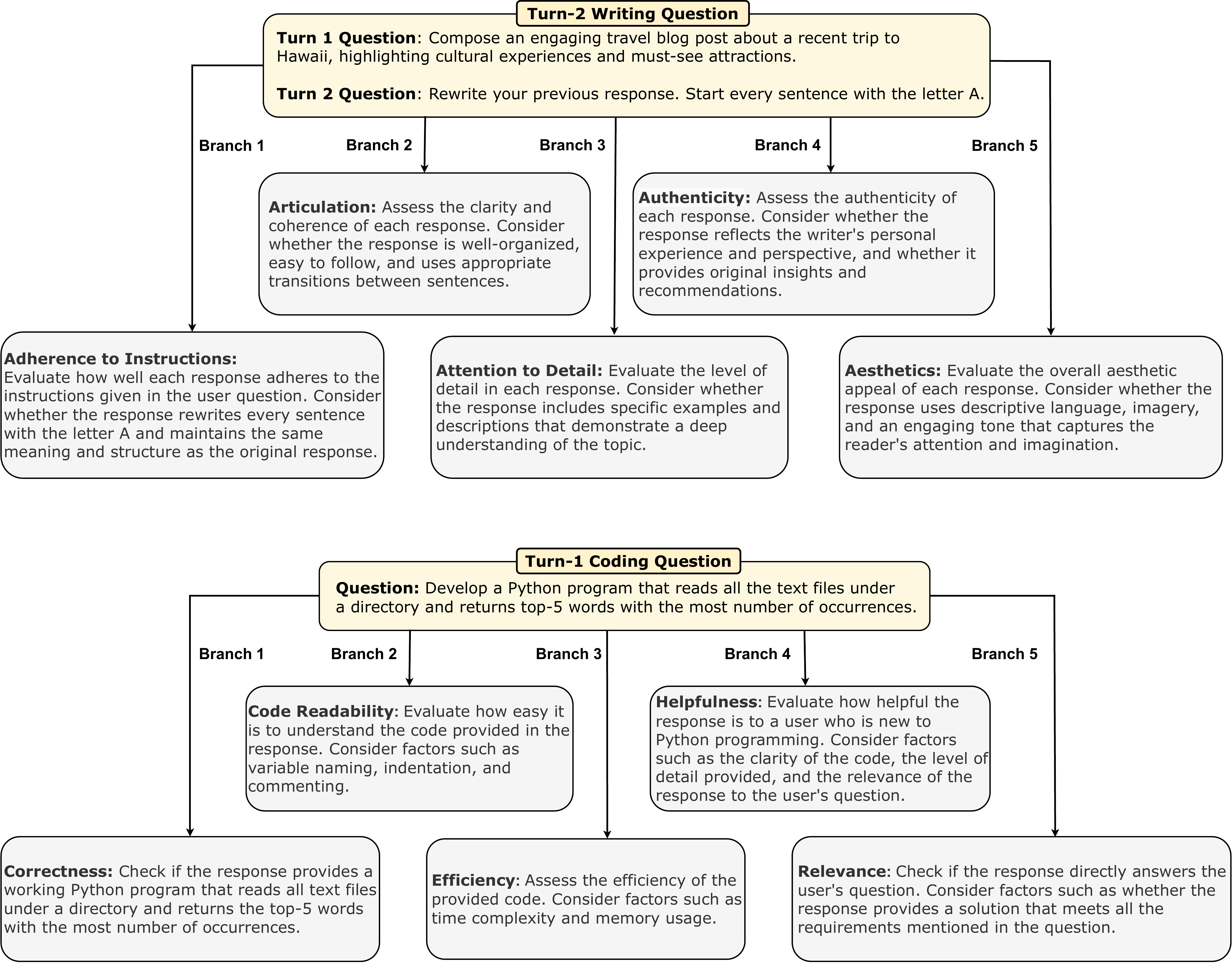}
      \vspace{-10pt}
  \caption{Examples of LLM Evaluation branch generation. We show different branches (evaluation plans) generated by BSM with a LLaMA-2-70B-chat model for different kinds of questions: (top) a turn-2 writing question and (bottom) a coding question. 
  }
  \label{fig:branch_turn2_code}
\end{figure*}

\begin{figure*}[t]
    \centering
    \includegraphics[width=0.9\linewidth]{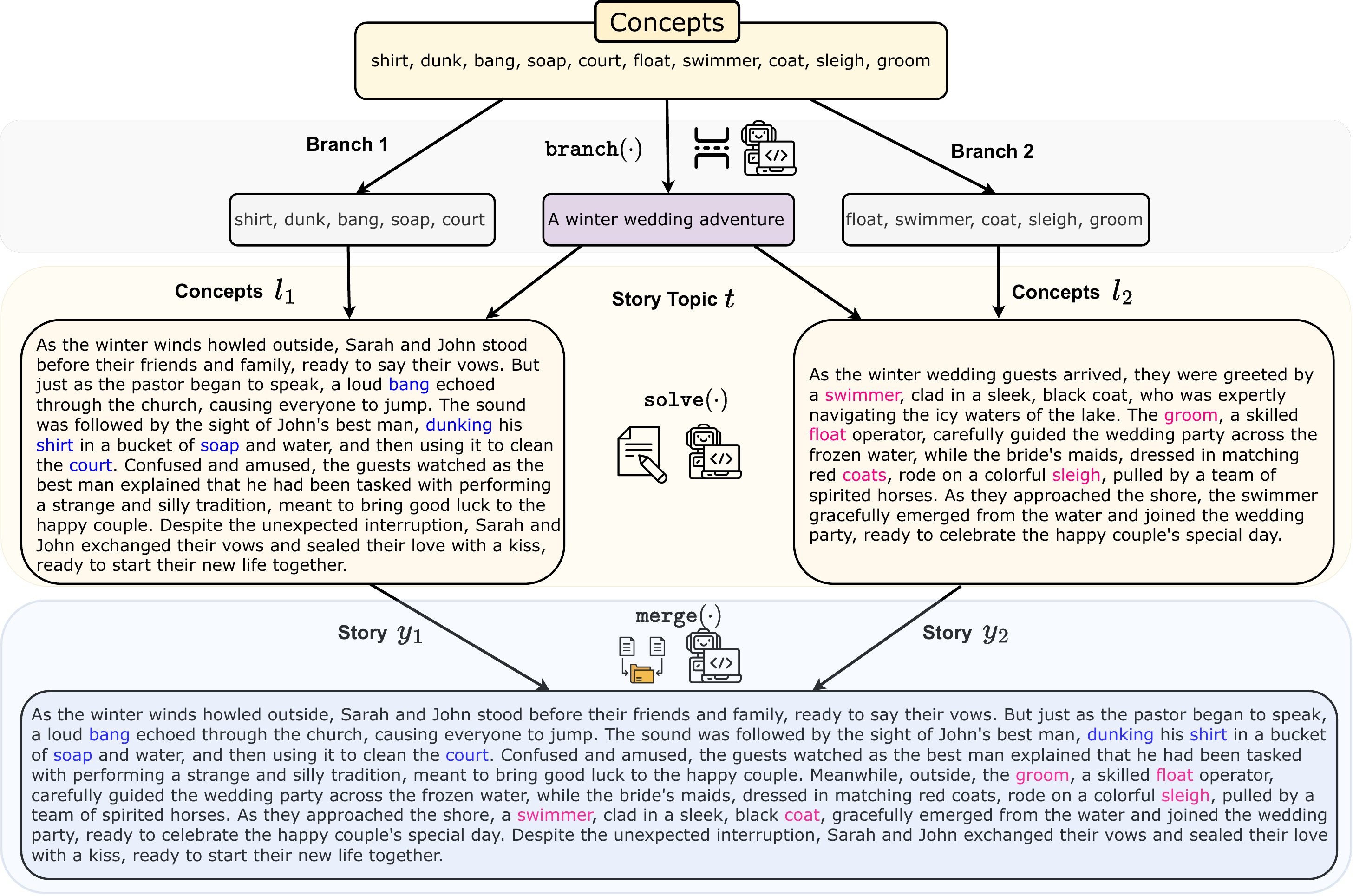}
    \vspace{-5pt}
    \caption{An illustration of \methodsmall{} with LLaMA-2-70B-chat for constrained story generation. Given a set of random concepts, the `Branch' module first divides them into two sets and generates a story topic. The `Solve' module conditions on the concepts and the topic to generate an intermediate story for each branch. The `Merge' module merges the intermediate stories to generate a final story ensuring that all concepts are still present.}
        \vspace{-2pt}
    \label{fig:constrained_gen}
\end{figure*}
\

\paragraph{Implementation Details.} Algorithm~\ref{algo:eval} shows the LLM program. For better reproducibility, all modules in BSM generate text using greedy decoding. For the branch module, the LLM is prompted to generate a plan consisting of a maximum of five evaluation criteria (which we found it adheres to in experiments). For the merge module, we find that the non-neural merge of summing up the criterion-wise evaluations is simple and works well in practice, hence all our experimental results are reported with that method. The prompts are shown in Figures~\ref{fig:branch_prompt} and~\ref{fig:solve_prompt}. All experiments are run on an AWS cluster of 8 A100 GPUs.

\paragraph{Baselines.} All methods, including BSM, account for position bias in the same manner, generating a verdict for both encoding orders and choosing the final verdict based on the individual verdicts (assigning a tie if the two encoding orders disagree). In particular, Self-Consistency computes majority vote independently for each encoding order.

\subsection{Results and Analysis}
\label{appendix:eval_results}
\paragraph{\methodsmall{} generalizes well across domains.}
In Table~\ref{tab:eval_no_reference}, we evaluate BSM's ability to evaluate generations for questions in the categories of `Roleplay', `Extraction', `Stem', and `Humanities'. We find that BSM is robust and performs well across domains in terms of improvement over the LLaMa-2-70B-chat baseline, and approaches GPT-4 performance on several of the domains. In particular, on the Stem domain, it is able to improve agreement scores over the baseline by up to 26\% (absolute), match GPT-4, and even outperform it in terms of position and length biases. Table~\ref{tab:reasoning_without_ref} shows that BSM outperforms the zero-shot baseline for `Reasoning' questions, even in reference-free evaluations (i.e., GPT-4 generated answers are not used either in the baseline or in BSM).

\paragraph{Combining BSM and SC reduces position bias further.} BSM generates a single solution for each sub-problem (each branch). A possible enhancement is combining BSM with self-consistency i.e., sampling multiple solutions for each sub-problem. In particular, we implement BSM+SC by sampling five evaluations per branch (with temperature $0.7$) and then the score for each sub-evaluation in that branch is given by the average score. We compare BSM with BSM+SC in Table~\ref{tab:bsm_sc}. While agreement scores do not improve further, we observe a 2\% reduction in position bias. This points to two conclusions. First, BSM, through its decomposition approach, already constructs sub-problems that are granular enough and hence, the variance reduction that one obtains through self-consistency within each sub-problem is limited. However, the moderate reduction in position bias still reflects its usefulness, which is a direct effect of making evaluations more consistent.

\begin{figure*}[!h]
    \centering
    \includegraphics[width=\textwidth]{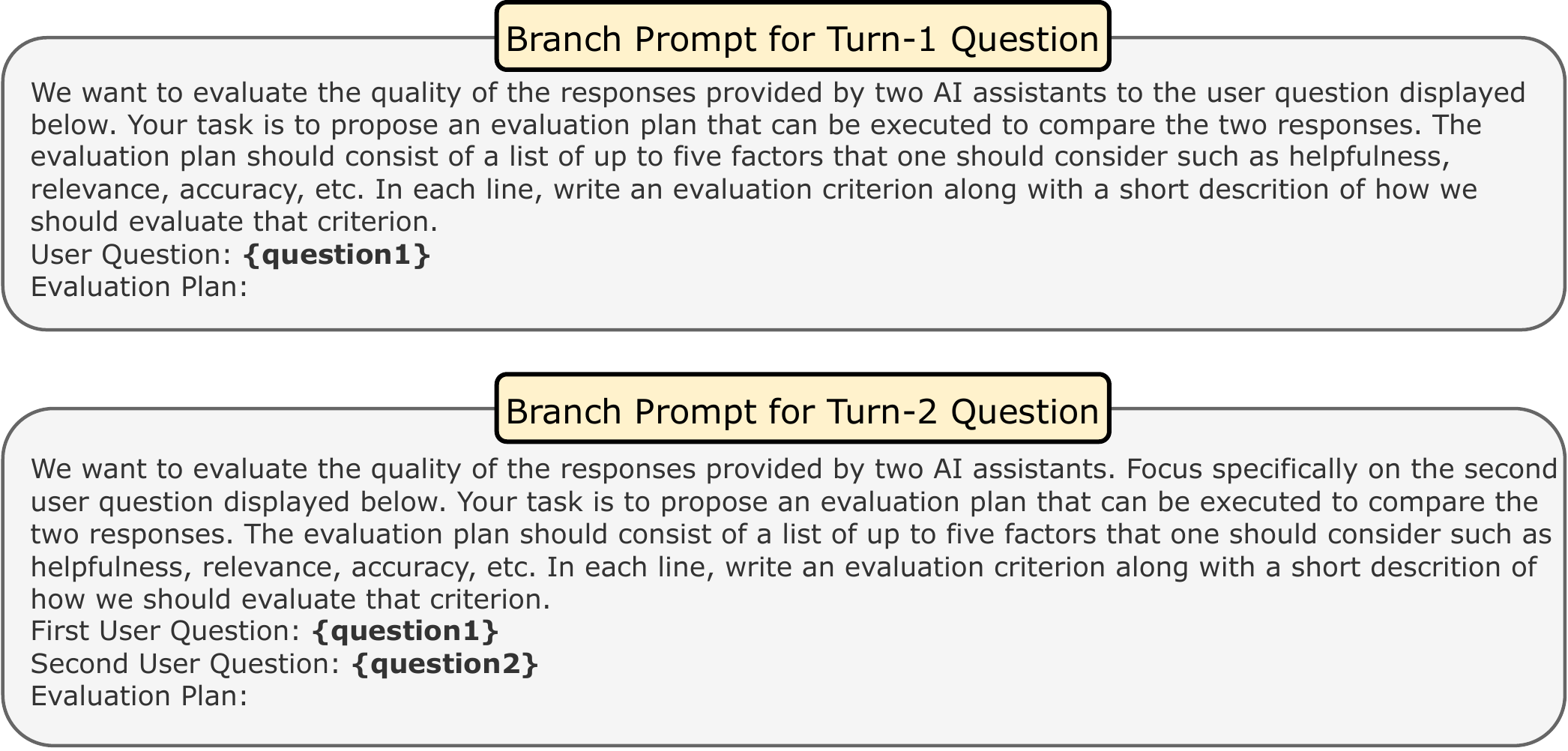}
    \vspace{-10pt}
    \caption{Branch prompts for Turn-1 and Turn-2 questions for the task of LLM Response Evaluation. The Turn-1 prompt conditions only on the Turn-1 question while the Turn-2 prompt conditions on both turn questions to generate an evaluation plan.}
        \vspace{-10pt}
    \label{fig:branch_prompt}
\end{figure*}

\paragraph{Effect of Branching Factor.} BSM has the benefit of relying on the underlying LLM for deciding what sub-problems to branch to, while the prompt controls the maximum branching factor (see the phrase `list of up to five factors' in the branch prompt in Fig.~\ref{fig:branch_prompt}). We vary this maximum branching factor from 2 to 5 and study its effect on 100 samples from the `writing' category of questions. Table~\ref{tab:branch} reports our findings. We observe highest agreement at a branching factor of 4, after which the result mostly saturates. In general, the optimal branching factor should depend on the specific question under consideration and unlike past work where users specify what factors to evaluate on~\citep{liu2023gpteval, zheng2023judging}, BSM generates that plan on its own. Position bias continues to decrease with increasing branching factor, where more branches helps reduce variance in the final judgment.

\begin{figure*}[!h]
    \centering
    \includegraphics[width=\textwidth]{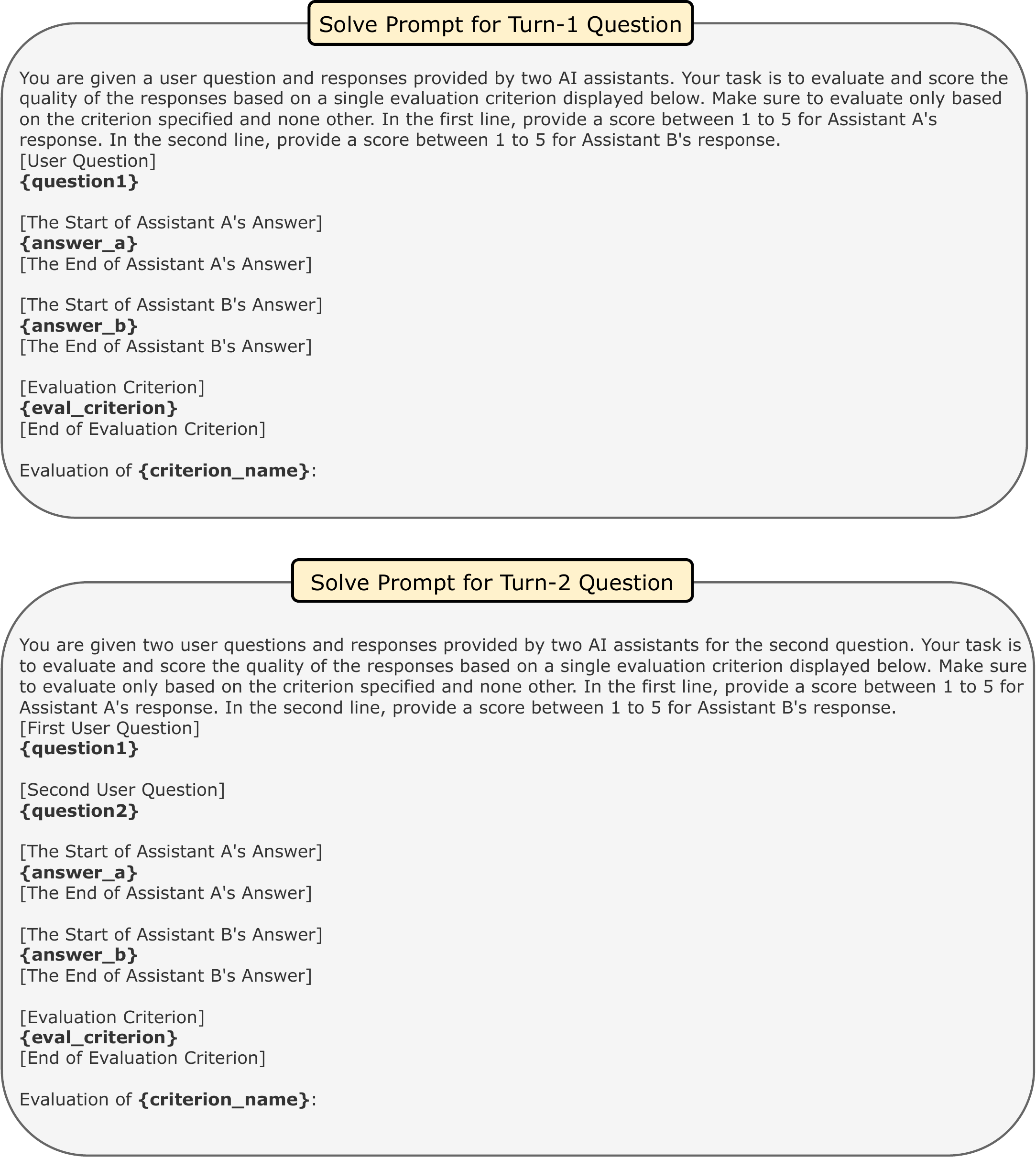}
    \vspace{-10pt}
    \caption{Solve prompts for Turn-1 and Turn-2 questions for the task of LLM Response Evaluation. Each prompt conditions on the question(s), the responses from both LLMs, and a given evaluation criterion that the branch module has generated.}
        \vspace{-5pt}
    \label{fig:solve_prompt}
\end{figure*}

\paragraph{BSM is robust to evaluation scale.} Evaluation tasks, in general, require defining a scale for scoring the responses. In Table~\ref{tab:solve}, we compare the performance of BSM by varying this evaluation scale, specified in the `solve' prompt (see Fig~\ref{fig:solve_prompt}), either scoring 1-5 (used in the main experiments) or 1-10. We observe that \methodsmall{} is fairly robust to such variations, obtaining comparable agreement scores. The position bias, however, increases slightly with a larger scale.

\section{Additional Experiments: Constrained Text Generation} 
\label{appendix:story_gen}

\paragraph{Analysis of Missing Concepts in BSM.} The source of missing concepts in BSM can be attributed to one of the following two categories: (a) the `solve' module, i.e., the model omits concepts even when generating an intermediate story in a branch subproblem with a lesser number of concepts; or (b) the `merge' module, i.e., the intermediate stories include their respective concepts but the fusion process omits some of these. We observe that out of 72\% of the BSM stories (with LLaMA-2-70B-chat) where at least one concept is missing, a significant 60\% of these belong to the first category (i.e., concept omission in the `solve' module) versus only 12\% belong to the second category (i.e., concept omission during `merging'). This suggests that constraint satisfaction can be further improved via a `recursive' BSM method involving iterative branching to even more granular sub-problems. However, recursive BSM would be significantly more expensive because of many more calls to the base LLM. We leave this exploration as part of future work.

\begin{algorithm*}[t]
\small
\caption{\method{}: LLM Program for LLM Evaluation}\label{algo:eval}
\begin{algorithmic}
\Require Question $q$, LLM Responses \{$r^{(A)}, r^{(B)}\}$, Modules $m$ = \{\texttt{branch($\cdot$),solve($\cdot$),merge($\cdot$)}\}
\Function{\textproc{BSM}}{$q, r^{(A)}, r^{(B)}, m, \mathit{swap}$}
\State {$C \gets \texttt{branch}(q)$} \Comment{Branch to different evaluation criteria.}
\ForEach {$c_i \in C$}
\If {$\mathit{swap} = \mathit{False}$}
\State {$s_i^{(A)}, s_i^{(B)} \gets \texttt{solve}(q, r^{(A)}, r^{(B)}, c_i)$} \Comment{Solve for each of the criteria.}
\Else
\State {$s_i^{(A)}, s_i^{(B)} \gets \texttt{solve}(q, r^{(B)}, r^{(A)}, c_i)$}
\EndIf
\EndFor
\State {$y^{(A, B)} \gets \texttt{merge}(q, \{c_i\}_{i=1}^k, \{s_i^{(A)}\}_{i=1}^k,  \{s_i^{(B)}\}_{i=1}^k) $} \Comment{Merge the individual evaluations.}
\State \Return $y^{(A, B)}$
\EndFunction

\Function{\method{}}{$q, r^{(A)}, r^{(B)}, m$}
\State $y^{(A, B)} = \textproc{BSM}(q, r^{(A)}, r^{(B)}, m, \mathit{swap} = \mathit{False})$ \Comment{Get verdict by not swapping the response order.}
\State $y^{(B, A)} = \textproc{BSM}(q, r^{(A)}, r^{(B)}, m, \mathit{swap} = \mathit{True})$ \Comment{Get verdict by swapping the response order.}
\If {$y^{(A, B)} = A \And y^{(B, A)} = B$}
\State $y \gets A$ \Comment{Choose a response only if the individual evaluations are consistent.}
\ElsIf {$y^{(A, B)} = B \And y^{(B, A)} = A$}
\State $y \gets B$
\Else 
\State $y \gets \mathit{tie}$
\EndIf
\State \Return $y$
\EndFunction
\end{algorithmic}
\label{alg:eval_bsm}
\end{algorithm*}

\begin{figure*}[!h]
    \centering
    \includegraphics[width=\textwidth]{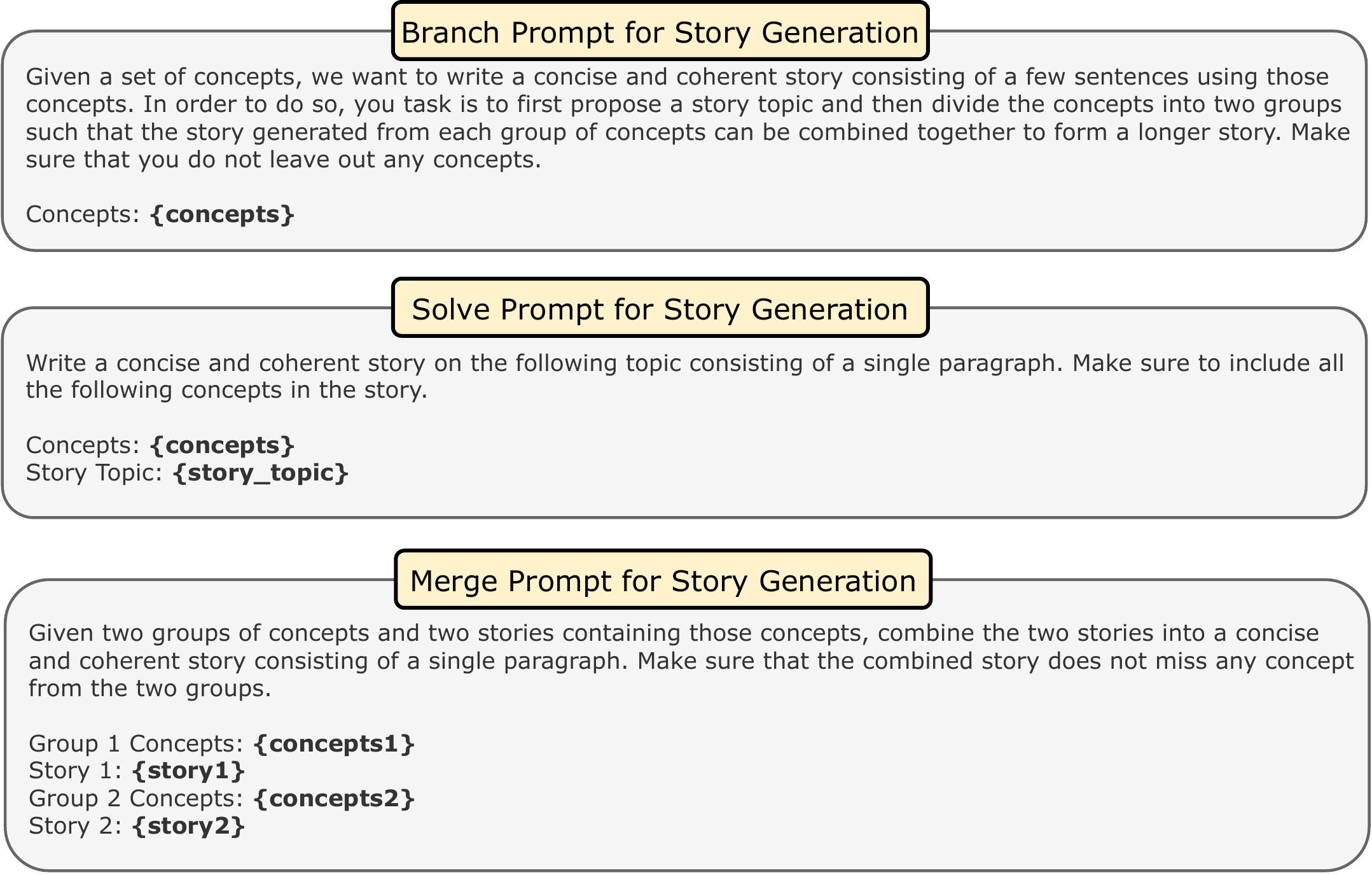}
    \vspace{-10pt}
    \caption{Branch, Solve, and Merge prompts for the task of Constrained Story Generation. The branch prompt conditions on the concepts. The solve prompt conditions on a subset of concepts and a story topic that the branch module has generated. The merge prompt conditions on the two intermediate stories that the solve module has generated and their respective concept-sets.}
        \vspace{-10pt}
    \label{fig:story_gen_all_prompts}
\end{figure*}

\begin{figure*}[!h]
    \centering
    \includegraphics[width=\textwidth]{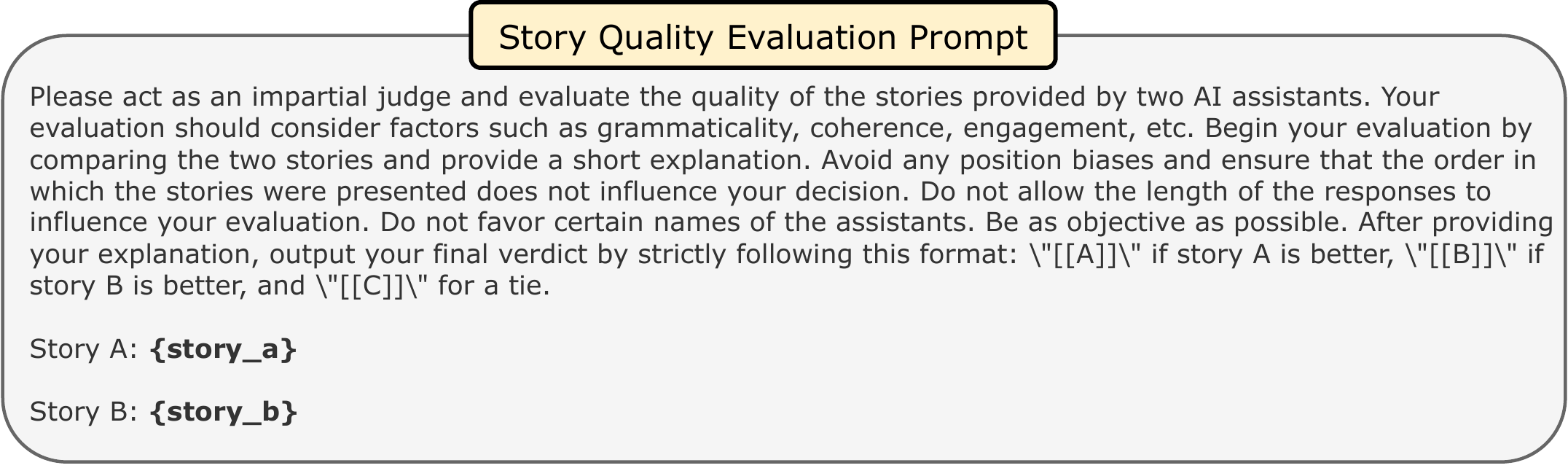}
    \vspace{-10pt}
    \caption{Prompt for evaluating the quality of the stories with GPT-4. It asks the model to perform a pair-wise evaluation between the stories generated by the baseline method and by BSM.}
        \vspace{-10pt}
    \label{fig:story_eval_prompt}
\end{figure*}

\end{document}